\UseRawInputEncoding
\pdfoutput=1

\documentclass[conference]{IEEEtran}
\IEEEoverridecommandlockouts
\usepackage{cite}
\usepackage{amsmath,amssymb,amsfonts}
\usepackage{algorithmic}
\usepackage{graphicx}
\usepackage{textcomp}
\usepackage{xcolor}
\usepackage{makecell}
\usepackage{caption}
\usepackage{bbding}
\usepackage{subcaption}
\usepackage{booktabs}
\usepackage{subcaption}
\usepackage{multirow}
\usepackage{threeparttable}
\usepackage{graphicx}
\usepackage{xspace}
\usepackage{xurl}
\usepackage{hyperref}
\newcommand{\ourmethod}{{PREM}\xspace}

\newcommand{\tabincell}[2]{\begin{tabular}{@{}#1@{}}#2\end{tabular}}  

\newcommand{\yx}[1]{\textcolor{black}{#1}}

\newcommand{\bill}[1]{\textcolor{black}{#1}}

\def\BibTeX{{\rm B\kern-.05em{\sc i\kern-.025em b}\kern-.08em
    T\kern-.1667em\lower.7ex\hbox{E}\kern-.125emX}}
\begin{document}

\title{PREM: A Simple Yet Effective Approach for Node-Level Graph Anomaly Detection
}
\renewcommand{\thefootnote}{\fnsymbol {footnote}}

\author{
\IEEEauthorblockN{Junjun Pan$^{*1}$, Yixin Liu$^{*2}$, Yizhen Zheng$^{*2}$, and Shirui Pan$^{1, 3}$\thanks{Shirui Pan is the corresponding author. }\thanks{* indicates equal contributions to this work. }}
\textit{$^1$ School of Information and Communication Technology, Griffith University, Queensland, Australia} \\
\textit{$^2$ Department of Data Science and AI, Monash University, Melbourne, Australia} \\
\textit{$^3$ Institute for Integrated and Intelligent Systems (IIIS), Griffith University, Queensland, Australia} \\
\IEEEauthorblockA{junjun.pan.joseph@outlook.com, \{yixin.liu, yizhen.zheng1\}@monash.edu, s.pan@griffith.edu.au}
}
\maketitle
\author{\IEEEauthorblockN{1\textsuperscript{st} Given Name Surname}
\IEEEauthorblockA{\textit{dept. name of organization (of Aff.)} \\
\textit{name of organization (of Aff.)}\\
City, Country \\
email address or ORCID}

}

\maketitle

\begin{abstract}
\bill{\yx{Node-level} graph anomaly detection (GAD) plays a critical role in identifying \yx{anomalous nodes from graph-structured data} in various domains such as medicine, social networks, and e-commerce. However, challenges have arisen due to \yx{the diversity of anomalies and} the dearth of labeled data. Existing methodologies - reconstruction-based and contrastive learning - while effective, often suffer from efficiency issues, stemming from their complex objectives and elaborate modules. To improve the efficiency of GAD, we introduce \yx{a simple method termed \underline{PRE}processing and \underline{M}atching} (\ourmethod \yx{for short}). Our approach streamlines GAD, \yx{reducing time and memory consumption while maintaining powerful anomaly detection capabilities.} Comprising two modules - a pre-processing module and an ego-neighbor matching module - \ourmethod eliminates the necessity for message-passing propagation during training, and employs a simple contrastive loss, leading to considerable reductions in training time and memory usage. Moreover, through rigorous evaluations of five real-world datasets, our method demonstrated robustness and effectiveness. Notably, when validated on the ACM dataset, \ourmethod achieved a 5\% improvement in AUC, a 9-fold increase in training speed, and sharply reduce memory usage compared to the most efficient baseline.}
\end{abstract}



\section{Introduction}






\bill{Graph anomaly detection (GAD) is a burgeoning field that focuses on identifying unusual patterns, including anomalous nodes, edges, or graphs that deviate noticeably from the majority~\cite{ding2019deep, liu2021anomaly, zhou2021subtractive}. 
\yx{As an advanced and practical technique, GAD has been} widely used in fields as diverse as medicine~\cite{mao2019medgcn, liu2023good}, social networks~\cite{li2021relevance}, and e-commerce~\cite{xu2019relation}.} 
\yx{Among various GAD scenarios, node-level GAD holds significant importance as it focuses on discerning the abnormality of individual nodes. }
\yx{For instance, node-level GAD can identify fraudsters in e-commerce networks, thereby contributing to improved security and trust within the online marketplace.}
\yx{In this paper, our research is specifically dedicated to addressing the challenging yet practical research problem of node-level GAD\footnote{In the paper, the term ``GAD'' specifically indicates node-level GAD.}.}

\yx{In practice, building a principled GAD model is a challenging task due to two major obstacles: the diversity of anomalies and the absence of anomaly labels~\cite{ding2019deep, liu2021anomaly}. Firstly, anomalous nodes in graphs can manifest in various ways, including distinctive node features from their neighborhoods, atypical interaction patterns with other nodes, and unusual topological substructures~\cite{ding2019deep}.  Consequently, the challenge lies in developing a single GAD model that can effectively identify diverse types of anomalous nodes comprehensively. On the other hand, the expensive nature of annotating anomaly labels makes it difficult to train a GAD model with reliable supervision signals~\cite{jin2021anemone}. As a result, traditional shallow (graph) anomaly detection methods often struggle to achieve satisfactory performance in such scenarios~\cite{peng2018anomalous, breunig2000lof, li2017radar, perozzi2016scalable}.} 

\begin{figure}
\centering
\begin{subfigure}{.25\textwidth}
  \centering
  \includegraphics[width=0.92\linewidth]{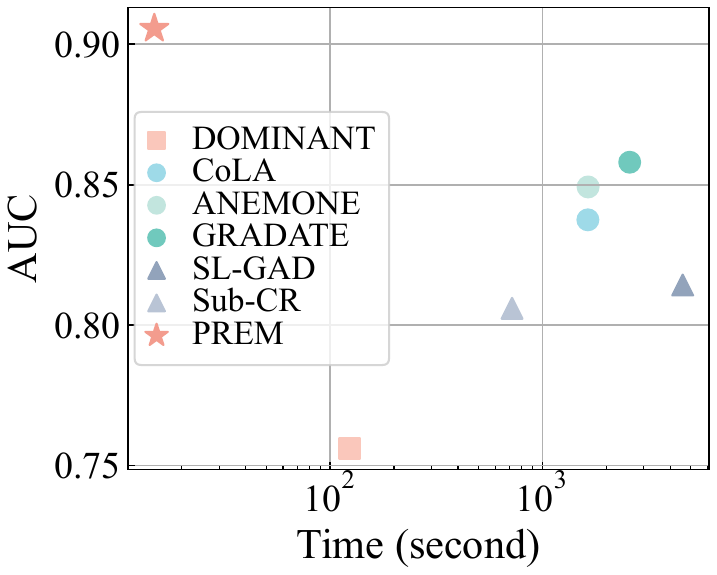}
  \caption{Train}

\end{subfigure}%
\begin{subfigure}{.25\textwidth}
  \centering
  \includegraphics[width=0.92\linewidth]{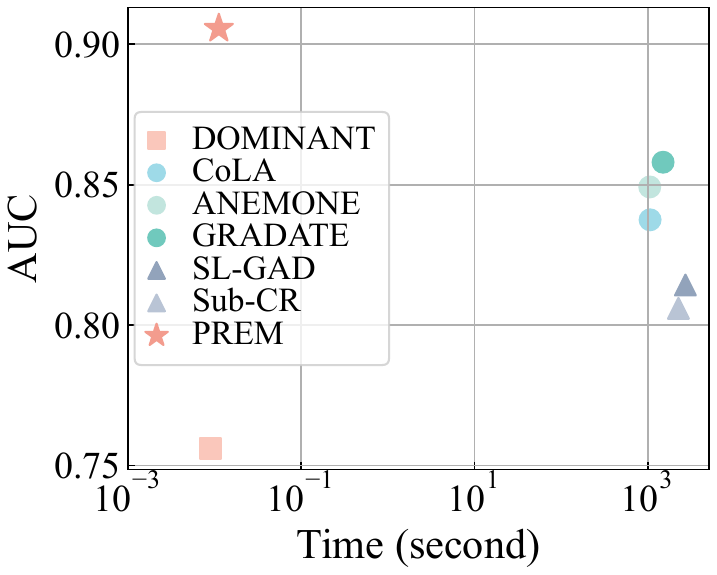}
  \caption{Test}
\end{subfigure}
\caption{AUC and runtime (in second) of baselines and our method \ourmethod on ACM. Left: training; Right: testing. }
\vspace{-3mm}
\label{fig:intro1}
\end{figure}


\yx{With the rising popularity of graph neural networks (GNNs), which have shown success in various graph-related tasks, there has been an emergence of deep learning-based methods proposed to tackle the GAD problem. These methods leverage unsupervised GNN models to capture intricate graph anomaly patterns and learn meaningful representations for anomaly detection purposes. Essentially, deep GAD methods can be categorized into two main categories: generative and contrastive. The generative GAD methods mainly use auto-encoder-like GNN models for graph data reconstruction and calculate anomaly scores based on the reconstruction error of nodes~\cite{ding2019deep, fan2020anomalydae, li2019specae, luo2022comga, zhang2022unsupervised}. The contrastive GAD methods, differently, employ a discriminator to assess the degree of inconsistency between target nodes and their neighbors for abnormality evaluation~\cite{liu2021anomaly, jin2021anemone, duan2022graph}. } 

\bill{Though \yx{the deep learning-based} methods have led to significant improvements for GAD, these advancements have come at a considerable cost in terms of efficiency in time and memory consumption. This is because these methods often rely on well-hand-crafted modules and complicated objectives to achieve great performance. For example, the multiple rounds estimation used in contrastive GAD prolongs the model evaluation time. In addition, the integration of structural-aware modules~\cite{zhang2022unsupervised, luo2022comga} or edge reconstruction tasks~\cite{ding2019deep, fan2020anomalydae, zhang2022unsupervised} intensifies memory usage, which escalates with dataset size, impeding memory efficiency. These limitations hinder the application of these methods in real-world scenarios characterized by vast amounts of data, underscoring the need for more efficient GAD solutions.}

\begin{figure}
\centering
\includegraphics[width=.9\linewidth]{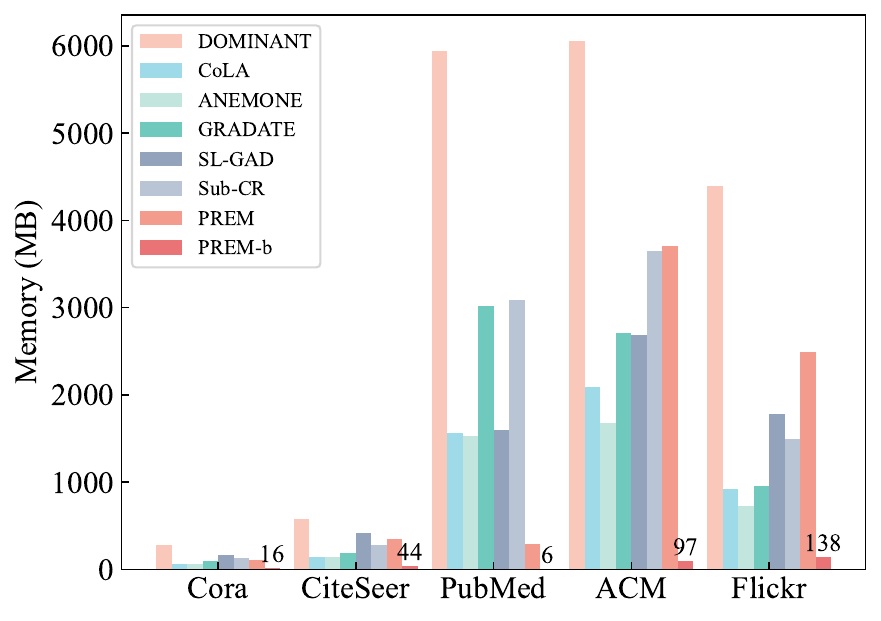}
\caption{GPU Memory \yx{usage (in MB) of} baselines and our method. Notices that we fix batch size=300 for all contrastive learning-based methods and \yx{the mini-batch version of \ourmethod (termed \ourmethod-b)} for comparison purposes. }
\label{fig:intro2}
\end{figure}

\bill{\yx{To enable effective, efficient, and computationally friendly anomaly detection on real-world graph data,} in this paper, we introduce a novel \yx{GAD method termed \underline{PRE}processing and \underline{M}atching}, herein referred to as \ourmethod. This framework offers superior performance over existing methods, coupled with exceptional efficiency in both time and memory. For instance, when evaluated on the ACM dataset, as depicted in Fig.~\ref{fig:intro1} and Fig.~\ref{fig:intro2}, our method substantially surpasses all existing GAD baselines. It registers an impressive performance improvement of 5\% in AUC, while significantly reducing the training time (9 $\times$ faster) and consumes much less memory compared with the most efficient baseline. The crux of \ourmethod lies in its streamlined two-part structure: a pre-processing module and an ego-neighbor matching module. Specifically, our approach eliminates message passing propagation during training, as the pre-processing stage aggregates neighbor features only once, leading to much lower time and space complexity. In addition, we employ a similarity-based method, eschewing the need for multi-round procedures that usually hamper efficiency. Furthermore, with the implementation of mini-batch processing, \ourmethod requires only a minimal amount of memory. In summary, the contribution of our work is three-fold:}
\bill{
\begin{itemize}
    \item We develop a simple yet very efficient framework, \underline{PRE}processing and \underline{M}atching, \ourmethod, for graph anomaly detection (GAD), adhering to the principle of simplicity and minimalism.
    \item Our proposed method, namely \ourmethod is orders of magnitude faster and consumes much less memory compared with existing GAD baselines.
    \item The robustness and effectiveness of our method have been thoroughly evaluated and analyzed across five real-world datasets, where \ourmethod achieved the best performance compared with existing GAD baselines.
\end{itemize}}




\begin{table*}[h!]
\centering
\caption{\bill{Architecture} comparison between state-of-the-art GAD methods and \ourmethod.}
\label{table:baselines}
\begin{threeparttable}
\begin{tabular}{c | p{1.8cm}<{\centering} p{1.8cm}<{\centering} p{1.8cm}<{\centering} p{1.8cm}<{\centering} p{1.8cm}<{\centering} p{1.8cm}<{\centering} p{1.8cm}<{\centering} } 
 \toprule
 Method & \tabincell{c}{Feature\\Reconstruction} & \tabincell{c}{Structure\\Reconstruction} & \tabincell{c}{Subgraph\\Sampling} & \tabincell{c}{Node-Subgraph\\Contrast}  & \tabincell{c}{Node-Node\\Contrast} & \tabincell{c}{Multi-View\\Contrast} & \tabincell{c}{Multi-round\\Evaluation} \\
 \midrule
 DOMINANT & \CheckmarkBold & \CheckmarkBold & \XSolidBrush & \XSolidBrush & \XSolidBrush & \XSolidBrush & \XSolidBrush\\
 CoLA & \XSolidBrush & \XSolidBrush & \CheckmarkBold & \CheckmarkBold & \XSolidBrush & \XSolidBrush & \CheckmarkBold\\
 ANEMONE & \XSolidBrush & \XSolidBrush & \CheckmarkBold & \CheckmarkBold & \CheckmarkBold & \XSolidBrush & \CheckmarkBold\\
 GADMSL & \XSolidBrush & \XSolidBrush & \CheckmarkBold & \CheckmarkBold & \CheckmarkBold & \CheckmarkBold & \CheckmarkBold\\
 SL-GAD  & \CheckmarkBold & \XSolidBrush & \CheckmarkBold & \CheckmarkBold & \XSolidBrush& \CheckmarkBold & \CheckmarkBold \\
 Sub-CR & \CheckmarkBold & \XSolidBrush & \CheckmarkBold & \CheckmarkBold & \XSolidBrush& \CheckmarkBold & \CheckmarkBold \\
 \bottomrule
 \ourmethod & \XSolidBrush & \XSolidBrush & \XSolidBrush & \CheckmarkBold\tnote{*} & \CheckmarkBold\tnote{*} & \XSolidBrush & \XSolidBrush\\
 \bottomrule
\end{tabular}

\begin{tablenotes}
\item[*] Without training-time propagation
\end{tablenotes}
\end{threeparttable}
\vspace{-3mm}
\end{table*}

\section{Related Work}

\subsection{Graph Neural Networks}

\bill{
Graph Neural Networks (GNNs), falling under the broad umbrella of deep neural network~\cite{lecun2015deep}, are specifically for handling semi-structured graph data. It is facilitated by harnessing both the attributive and the topological information inherently present within the non-Euclidean graph-structured data. The very inception of GNNs can be traced back to~\cite{bruna2013spectral}, where an innovative spectral-based method was introduced. This method facilitated the extension of convolution networks to graph-based structures. Following this inaugural contribution, the concept was further built upon with the development of a range of subsequent spectral-based convolution GNNs~\cite{defferrard2016convolutional,henaff2015deep}. These furthered the field by adopting filters designed in accordance with the principles of graph signal processing~\cite{shuman2013emerging, zheng2023finding}. One important milestone in this domain was the advent of Graph Convolutional Networks (GCN)~\cite{kipf2016semi}. The GCN essentially served as a bridge, connecting spectral-based methods and spatial-based GNN approaches by simplifying spectral graph convolutions. Specifically, it approximates the first-order Chebyshev polynomial filters.
}

\bill{
The development of GCN paved the way for the rapid evolution of spatial-based methods. These methods were not only more efficient but also exhibited wider applicability across various use cases. For example, Graph Attention Networks (GAT)~\cite{velivckovic2017graph} was a significant evolution in this regard. GAT integrated the attention mechanism~\cite{vaswani2017attention}, making it possible to account for varying levels of significance among node neighbors, rather than just employing a basic average of neighboring data. Another significant spatial-based approach was the Simplified Graph Convolutions (SGC)~\cite{wu2019simplifying}. It further simplifies GCN by removing non-linearity and collapsing weight matrices between graph convolution layers, reducing the overall complexity. In addition to these major contributions, numerous studies have focused on improving GNN from other unique angles. These include the expansion of scalability~\cite{bojchevski2020scaling, zeng2019graphsaint, zheng2022rethinking}, the reduction of prior knowledge requirement~\cite{liu2023learning, liu2023beyond, zhu2021graph, zheng2022unifying, zheng2022toward, zheng2023towards, zheng2023structure} and the enlargement of receptive fields~\cite{klicpera2019diffusion, tang2020cgd, liao2019lanczosnet,wu2021learning, zheng2022unifying}.}
\bill{
GNN approaches have found widespread success across multiple domains. They have been employed effectively in areas such as business analysis~\cite{yizhen2021hetergraph}, federated learning~\cite{tan2023federated}, latent structure inference~\cite{liu2022towards}, and question answering~\cite{luo_rog}.
}

\subsection{Traditional Graph Anomaly Detection}

Graph anomaly detection can be defined as classifying outliers from normal nodes. With the increasing use of attributed networks in industry, the importance of graph anomaly detection has become popular in recent years. 
\bill{In order to effectively detect anomalies in attributed networks, various non deep learning-based algorithms and methodologies have been developed, each with their own strengths and weaknesses.} A pioneer work,
AMEN~\cite{perozzi2016scalable} mines anomaly information from the internal and external consistency of neighborhoods. Radar~\cite{li2017radar} analyses the residual between target node attributes and the majority to calculate the anomaly score. ANOMALOUS\cite{peng2018anomalous} extends the framework of~\cite{li2017radar} by incorporating CUR decomposition with residual analysis. DEEPFD~\cite{wang2018deep} first attempts to learn the embedding via reconstructing target node's features, and then detects anomaly substructures via clustering.~\cite{zhu2020mixedad} improves the scalability by growing and merging communities of nodes that share similar properties. HCM~\cite{huang2022hop} learns the embedding by predicting the shortest distance between nodes to obtain both local and global information. \yx{However, due to the limitations of their shallow mechanism, their performance is proven to be limited when handling high-dimensional features and complex structures.}


\subsection{Deep Learning-based Graph Anomaly Detection}

\bill{\yx{Compared to traditional methods, Deep learning-based graph anomaly detection (GAD) methods are able to better capture the complex relationships and structures present in attributed networks, resulting in more accurate and efficient anomaly detection.} They mainly fall into two categories: generative and contrastive learning. DOMINANT~\cite{ding2019deep} pioneers the first \yx{generative GAD method} by learning node embeddings via minimizing reconstruction error for attributes and adjacency matrix, exploiting both attribute and topological features. SpecAE\cite{li2019specae} follows a similar approach but uses deconvolution and Gaussian Mixture Model for anomaly estimation. AnomalyDAE~\cite{fan2020anomalydae} decouples attribute and structural encoders to efficiently model interactions. ComGA~\cite{luo2022comga} incorporates community detection, while AS-GAE~\cite{zhang2022unsupervised} adds a location-aware module.}
\bill{For contrastive learning-based methods, CoLA~\cite{liu2021anomaly} addresses challenges with a contrastive learning paradigm, using a discriminator to detect inconsistencies between target node and neighbor subgraph embeddings. ANEMONE~\cite{jin2021anemone} introduces a patch-level contrastive task for multi-scale anomaly detection. GRADATE~\cite{duan2022graph} improves the framework through graph augmentation and multi-views contrast. SL-GAD~\cite{zheng2021generative} combines attribute reconstruction and node-subgraph contrast, while Sub-CR~\cite{zhang2022reconstruction} uses masked autoencoder and graph diffusion to fuse attributes with local and global topological information. \yx{Apart from the two main categories, There are also other methods targeting GAD, such as subtractive aggregation~\cite{zhou2021subtractive}, one-class SVM~\cite{wang2021one}, and complementary learning~\cite{liu2022dagad}.}}


\bill{
Although existing \yx{deep} GAD methods have demonstrated promising results, their limitations in efficiency and scalability hinder their applicability in real-world scenarios involving large-scale graphs with a vast number of nodes and edges. As shown in Table~\ref{table:baselines} and Fig.~\ref{fig:intro1},~\ref{fig:intro2}, integrating more modules improves performance but at the expense of efficiency. To address these issues, we propose \ourmethod, a novel approach that significantly streamlines the GAD process by employing a linear discriminator in an embarrassingly simple manner. Not only does \ourmethod outperform the baseline methods by a considerable margin, but it also runs much faster and has lower memory consumption. 
This enhanced efficiency and scalability make \ourmethod a more suitable choice for practical applications in detecting anomalies within extensive graph structures.}

\section{Preliminaries}

\subsection{Definition}
\yx{Throughout this paper, we use calligraphic fonts, bold lowercase letters, and bold uppercase letters to denote sets, vectors, and matrices, respectively.} 
We denote an attributed graph as $G=(\mathcal{V}, \mathcal{E}, \mathbf{X})$, where $\mathcal{V}=\{v_1, \cdots, v_n\}$ is the set of vertices (or nodes), the number of nodes $|\mathcal{V}|$ is $n$, and $\mathcal{E}$ is the set of edges, where $|\mathcal{E}|=m$, which can also be expressed by a binary adjacent matrix $\mathbf{A} \in \mathbb{R}^{n \times n}$. The feature matrix of a graph $\mathbf{X}  \in \mathbb{R}^{n \times d}$ can be represented as a list of row vectors $\mathbf{X} = (\mathbf{x}_1^T, \cdots, \mathbf{x}_n^T)$, where $\mathbf{x}_i$ is the feature vector of node $v_i$ and $d$ is the dimension of features.

\subsection{Problem Formulation}
\yx{In this paper, we focus on the unsupervised node-level graph anomaly detection (GAD) problem.} The aim is to learn a scoring function $f(\cdot)$ to estimate the anomaly score $s_i$ for each node $v_i$. A larger anomaly score $s_i$ indicates that the node $v_i$ is more likely to be an anomalous node. In this paper, we aim to develop a novel GAD model (serving as $f(\cdot)$) that estimates the anomaly score efficiently under strictly unsupervised settings without ground-truth labels of the anomaly provided. 

\section{Method}

\begin{figure*}
\centering
\includegraphics[width=.95\linewidth]{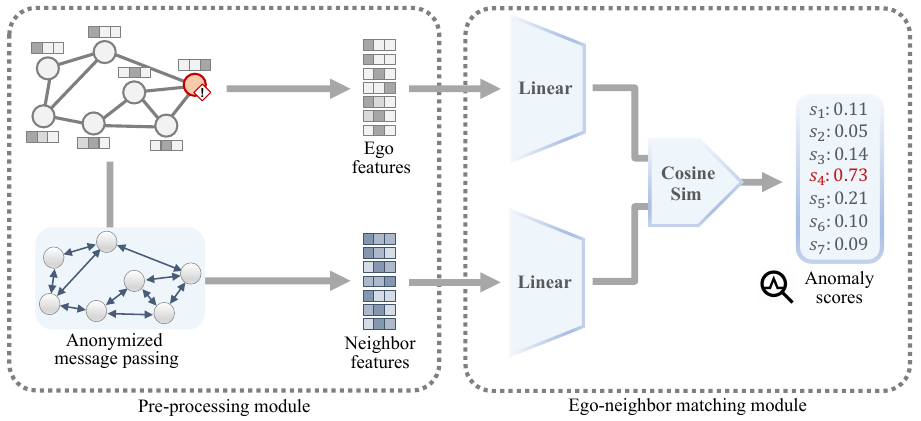}
\caption{The \yx{overall pipeline} of \yx{\ourmethod that composes of a pre-processing module and an ego-neighbor matching module}. \bill{In the first module, we conduct anonymized message passing, i.e., without the self-loop, on the given graph to obtain neighbor features. Then, in the ego-neighbor matching module, we input the ego features, i.e., raw features, and neighbor features into a discriminator, which includes a linear layer and the cosine similarity computation component, to obtain anomaly scores. The higher the score is, more likely the node is an anomaly.} }

\label{fig:Architecture}
\end{figure*}
\yx{In this section, we introduce the proposed model term \ourmethod in detail. The overall pipeline of \ourmethod is illustrated in Fig.~\ref{fig:Architecture}. On the highest level, \ourmethod is composed of two components:} \bill{1) A \textit{pre-processing module} is designed to precompute the neighbor features, thereby eliminating the need for costly message passing during training;}  \yx{and 2) a \textit{ego-neighbor matching module} that learns the matching patterns between ego and neighbor information with a simple linear layer-based network. Once \ourmethod is well-trained, the anomaly scores can be estimated by the ego-neighbor similarity directly. In the following subsections, we first introduce the two main components of \ourmethod respectively. Then, we illustrate a simple contrastive learning objective to train \ourmethod model. Finally, we expound on the calculation of anomaly scores during inference and discuss the superb efficiency of \ourmethod.}

\subsection{Message Passing-based Pre-Processing}

\yx{In conventional GAD methods, message passing-based GNNs are commonly used to capture contextual information of each node, aiding in the prediction of node-level abnormality.  However, the forward and backward propagation involved in message passing can be time-consuming, which significantly affects the efficiency of model training~\cite{han2023mlpinit}. Consequently, an important question arises: \textit{Is it possible to expedite the learning process by eliminating message passing during training?} To address this question, we propose a pre-processing scheme based on message passing that enables \ourmethod to perform message passing only once throughout the learning procedure.}

\yx{Since the agreement between each node and its context (i.e., its surrounding substructure) is highly correlated to the node-level abnormality~\cite{liu2021anomaly, jin2021anemone}, in our pre-processing module, we concentrate on extracting two types of information: ego information, which indicates the attributes specific to a node, and neighbor information, encompassing the contextual knowledge surrounding the node. Firstly, to capture ego information, we construct \textit{ego features} $\mathbf{X}^{(e)}$ by collecting the raw feature vectors together directly, i.e., $\mathbf{X}^{(e)}= \mathbf{X}$. This approach enables us to capture the intrinsic attributes and characteristics of each node without considering the surrounding context.}

\yx{Then, to extract neighbor information, we propose to conduct message passing on the raw features to generate \textit{neighbor features} $\mathbf{X}^{(n)}$. One possible approach is to adopt the propagation rule used in mainstream GNNs, such as GCN~\cite{kipf2016semi} and SGC~\cite{wu2019simplifying}, for our message passing-based pre-processing. 
However, it is important to note that conventional GNNs typically aggregate both ego and neighbor information during message passing. This aggregation strategy may lead to information leakage in the downstream ego-neighbor matching network, subsequently impacting the performance of anomaly detection. To alleviate this issue, we design an \textit{anonymized message passing scheme} that enables the extraction of neighbor information separately. In concrete, the neighbor features can be calculated by:}

\begin{equation}
\mathbf{X}^{(n)} = \mathbf{P} \mathbf{X}, 
\end{equation}

\noindent \yx{where $\mathbf{P}$ is the anonymized propagation matrix that can be written by:}

\begin{equation}
\mathbf{P} = \mathbf{M}*(\tilde{\mathbf{D}}^{-1/2}\tilde{\mathbf{A}}\tilde{\mathbf{D}}^{-1/2})^k,
\end{equation}

\noindent \yx{where $k$ is the propagation steps in message passing, $\tilde{\mathbf{A}}=\mathbf{A} + \mathbf{I}$ is the adjacency matrix with self-loop, $\tilde{\mathbf{D}}$ is the degree matrix of $\tilde{\mathbf{A}}$, and $\mathbf{M}$ is a self-anonymization mask where diagonal entries are all zero while the remaining entries are all one. By utilizing the self-anonymization mask, the signals associated with ego features can be filtered out during the message passing process. Consequently, each row vector $\mathbf{x}_i^{(n)}$ in the neighbor features matrix $\mathbf{X}^{(n)}$ can concentrate on summarizing the neighbor information independently, without being influenced by the ego information.}

\yx{It is worth noting that $\mathbf{X}^{(e)}$ and $\mathbf{X}^{(n)}$ are computed only once during the learning procedure, and there are no trainable weights to be involved in the pre-processing step. Thanks to this merit, our method is much simpler and faster than existing methods that involve multiple layers of GNNs in GAD models.} 

\subsection{Ego-Neighbor Matching Network} \label{subsec:matching}

\yx{Once we have obtained the ego and neighbor features, we proceed to construct an \textit{ego-neighbor matching network}. This network is designed to learn the matching patterns between the ego information and the corresponding neighbor information of each node. By capturing these matching patterns, we can effectively assess the normality of nodes during the testing phase.}

\yx{In specific, the proposed ego-neighbor matching network is composed of two linear layers and a cosine similarity unit. Note that the original features of attributed graphs are frequently high-dimensional, and there are often correlations among different feature dimensions. In such scenarios, directly measuring the similarity solely at the feature space can be inefficient and ineffective. Therefore, instead of calculating the feature-level similarity, we first introduce two learnable linear layers to map the ego features and neighbor features into a low-dimensional space, respectively. Formally, given the ego feature vector $\mathbf{x}^{(e)}_i$ and neighbor feature vector $\mathbf{x}^{(n)}_i$ for node $v_i$, the linear mapping can be written by:}

\begin{equation}
\begin{aligned}
\mathbf{h}^{(e)}_i &= \mathbf{x}^{(e)}_i \mathbf{W}_1 + \mathbf{b}_1, \\
\mathbf{h}^{(n)}_i &= \mathbf{x}^{(n)}_i \mathbf{W}_2 + \mathbf{b}_2 ,
\end{aligned}
\end{equation}

\noindent \yx{where $\mathbf{h}^{(e)}_i$ and $\mathbf{h}^{(n)}_i$ are low-dimensional ego embedding vector and neighbor embedding vector respectively, while $\mathbf{W}_1$, $\mathbf{W}_2$, $\mathbf{b}_1$, and $\mathbf{b}_2$ are learnable weights/bias parameters of the linear layers.}

\yx{After linear mapping, now we can evaluate the ego-neighbor similarity with the compact embeddings. To simplify, we employ a cosine similarity unit to measure the agreement between two embeddings, which can be formulated by:}

\begin{equation}
c_i = \operatorname{cos}(\mathbf{h}^{(e)}_i, \mathbf{h}^{(n)}_i),
\end{equation}

\noindent \yx{where $\operatorname{cos}(\cdot, \cdot)$ is the cosine similarity function and $c_i \in [-1,1]$ is the ego-neighbor similarity of node $v_i$. Intuitively, $c_i$ can be used to indicate the normality of node $v_i$: according to the homophily assumption, a (normal) node tends to have similar behavior with its neighborhoods, leading to a larger $c_i$ given by our matching network; in contrast, anomalous nodes usually break the homophily assumption due to their noisy connections and unexpected features~\cite{liu2021anomaly}, hence have smaller $c_i$. In other words, we can easily estimate the anomaly score according to the learned similarity. }



\begin{figure}
\centering
\includegraphics[width=.9\linewidth]{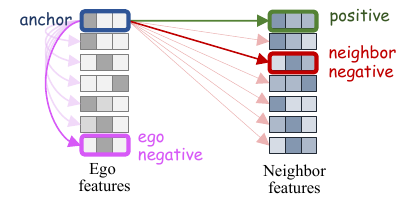}
\caption{Illustration of positive and negative pairs in \ourmethod.}

\label{fig:tasks}
\end{figure}

\subsection{Contrastive Learning-based Model Training}

\yx{In this subsection, we present the training strategy for our \ourmethod model. Given the challenge of obtaining ground-truth labels for anomalies, our focus in this paper is on the unsupervised GAD problem~\cite{ding2019deep, liu2021anomaly}. In this context, it becomes necessary to design a carefully crafted unsupervised learning objective to train the \ourmethod model, as opposed to relying on labeled data. Inspired by the success of contrastive learning on GAD tasks~\cite{liu2021anomaly, jin2021anemone}, we design a simple yet effective contrastive learning objective for our model training. }

\yx{Recalling that we aim to maximize the ego-neighbor similarity of a normal node. However, we cannot identify the normal nodes exactly without accessing their labels during the training phase. Fortunately, with the assumption that the number of anomalies is much smaller than which of normal nodes, we can directly maximize the ego-neighbor similarity of all nodes. By doing so, \ourmethod can effectively capture the dominant matching patterns observed in normal nodes. This approach allows us to prioritize the identification of normal behavior patterns while accommodating the inherent difficulty of precisely identifying individual normal nodes without explicit labels. By focusing on the majority of normal nodes, we can still train the model to capture the essential matching patterns associated with normalcy. Formatting this idea to a contrastive framework, we define the ego and neighbor embeddings of the same node, i.e., $\mathbf{h}^{(e)}_i$ and $\mathbf{h}^{(n)}_i$, as a positive pair. Then, their cosine similarity is denoted as a positive term in contrastive learning:}

\begin{equation}
c_i^{(pos)} = c_i = \operatorname{cos}(\mathbf{h}^{(e)}_i, \mathbf{h}^{(n)}_i).
\end{equation}

\yx{If we only maximize the positive term $c_i^{(pos)}$, it would lead to the model collapse problem, i.e., the model assigns high values or probabilities to all input pairs. The model collapse problem can severely reduce the discriminative power and hinder the effective detection of anomalies. To avoid this issue, it is necessary to involve negative pair in our contrastive training scheme. Regarding the ego embedding $\mathbf{h}^{(e)}_i$ of node $v_i$ as an anchor, we can naturally define the neighbor embedding $\mathbf{h}^{(n)}_j$ of another node $v_j$ as a negative sample. In this case, we can obtain a negative pair ($\mathbf{h}^{(e)}_i$ and $\mathbf{h}^{(n)}_j$) that forms a neighbor-based negative term in contrastive learning:}

\begin{equation}
c_i^{(neg_{nbr})} = \operatorname{cos}(\mathbf{h}^{(e)}_i, \mathbf{h}^{(n)}_j).
\end{equation}

\yx{With the neighbor-based negative term, now we can learn to discriminate the matching patterns between ego and neighbor information with \ourmethod. Nevertheless, it is important to note that relying solely on neighbor-based negative samples may introduce a potential issue known as the ``easy negative'' problem~\cite{zhu2019distance, robinson2021contrastive}. The easy negative problem occurs when the negative pairs provided by the neighbors are too straightforward for the model to discriminate, resulting in the model not effectively learning meaningful matching patterns. This can hinder the model's ability to accurately identify anomalies. To address the easy negative problem, we propose introducing more diverse and challenging negative samples. One effective strategy to achieve this is by leveraging ego features. Discriminating among ego features is particularly challenging because the raw features of two nodes can be very similar in a graph. Motivated by this, in addition to the neighbor-based negative term, we also incorporate an ego-based negative term for contrastive training. In concrete, given the ego embedding $\mathbf{h}^{(e)}_i$ as an anchor, we randomly sample the ego embedding $\mathbf{h}^{(e)}_k$ of another node $v_k$ as its ego-based negative sample. Then, the ego-based negative term can be written by:}

\begin{equation}
c_i^{(neg_{ego})} = \operatorname{cos}(\mathbf{h}^{(e)}_i, \mathbf{h}^{(e)}_k).
\end{equation}

\yx{Finally, we establish a contrastive loss function to maximize the positive contrastive terms while minimizing the negative contrastive terms. Unlike the majority of works~\cite{zhu2021graph, you2020graph, qiu2020gcc} that employ the Info-NCE contrastive loss, in \ourmethod, we use a binary cross-entropy-like contrastive loss for model training, which can be regarded as an extension of Jensen-Shannon divergence-based contrastive loss~\cite{velickovic2019deep, hassani2020contrastive}. Specifically, we first use a linear mapping to project the positive/negative terms $c \in [-1,1]$ into $\hat{c} \in [0,1]$, and then acquire the loss by:}

\begin{multline}
L = - \sum_{t=1}^{N} ( \log(\hat{c}_i^{(pos)}) + \alpha \log(1-\hat{c}_i^{(neg_{nei})}) + \\
\gamma \log(1-\hat{c}_i^{(neg_{ego})})),
\end{multline}
\noindent \yx{where $\alpha$ and $\gamma$ are the hyper-parameters for trade-off. Compared to Info-NCE loss with $\mathcal{O}(n^2)$ time complexity, our loss is much more efficient to compute (with $\mathcal{O}(n)$ complexity). Such a nice property ensures the rapid model training of \ourmethod. Fig.~\ref{fig:tasks} illustrates an example of the definitions of positive and negative pairs in \ourmethod.}

\begin{figure*}
\centering
\begin{minipage}[t]{0.3\linewidth}
\includegraphics[width=1\linewidth]{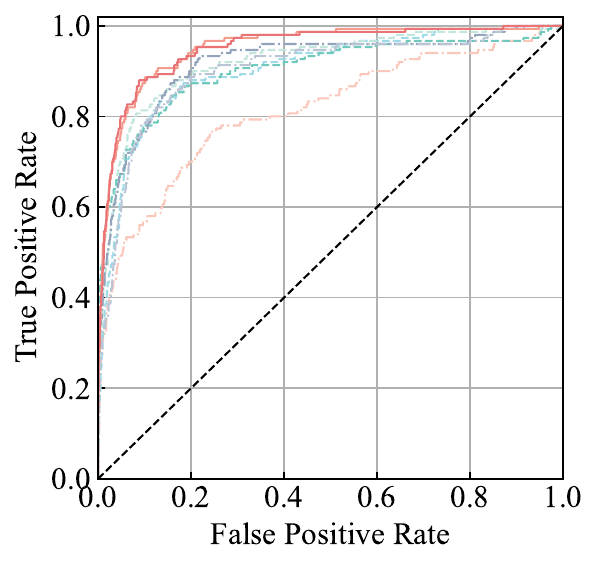}
\subcaption{Cora}
\end{minipage}
\begin{minipage}[t]{0.3\linewidth}
\includegraphics[width=1\linewidth]{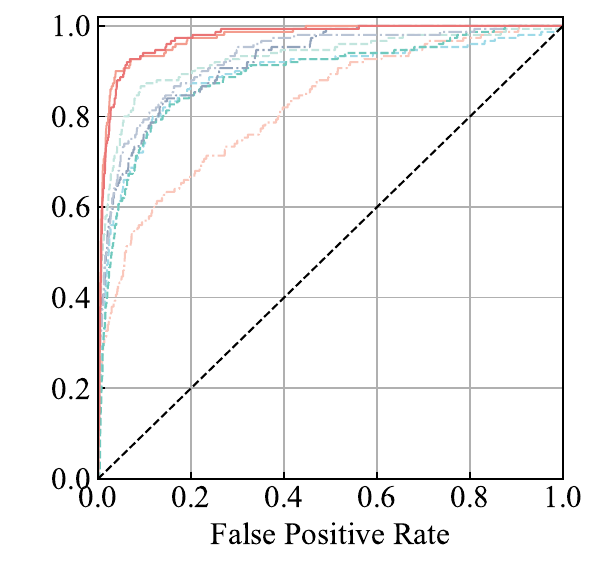}
\subcaption{CiteSeer}
\end{minipage}
\begin{minipage}[t]{0.3\linewidth}
\includegraphics[width=1\linewidth]{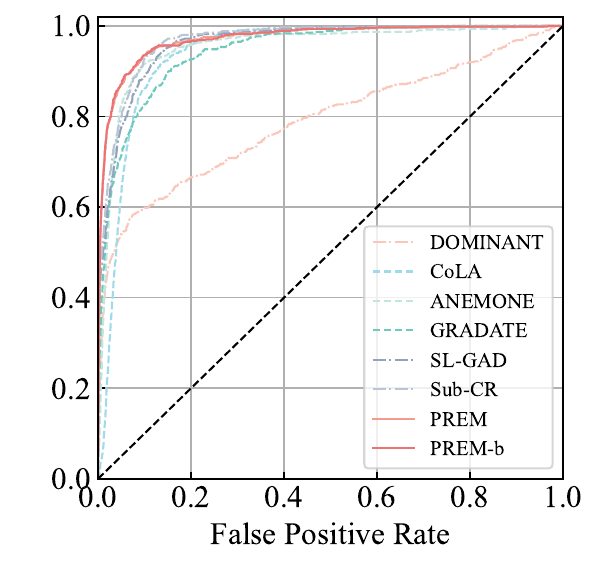}
\subcaption{PubMed}
\end{minipage}
\begin{minipage}[t]{0.3\linewidth}
\includegraphics[width=1\linewidth]{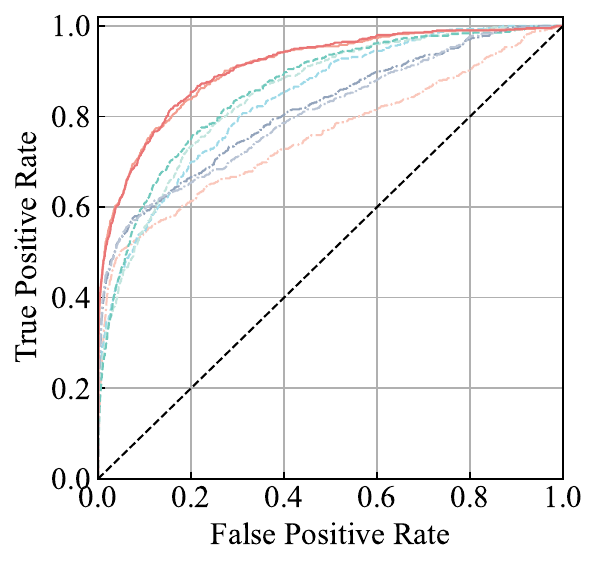}
\subcaption{ACM}
\end{minipage}
\begin{minipage}[t]{0.3\linewidth}
\includegraphics[width=1\linewidth]{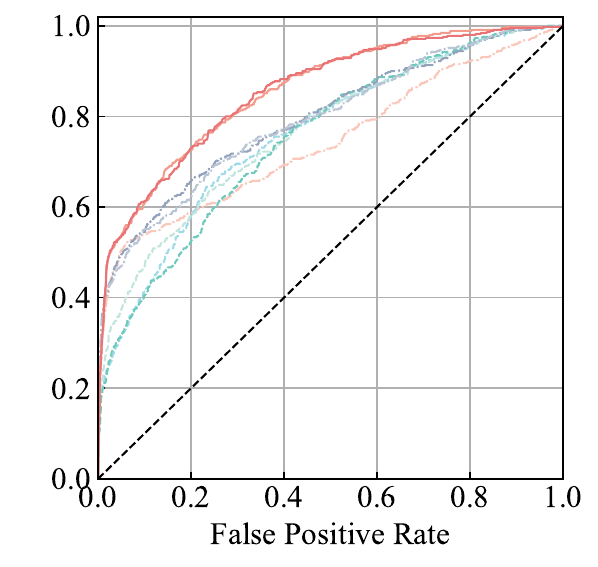}
\subcaption{Flickr}
\end{minipage}
\centering
\caption{ROC curves on five datasets.}

\label{fig:ROC}
\end{figure*}

\subsection{Anomaly Score Calculation}

\yx{Once the \ourmethod model is well-trained, we can estimate the anomaly score using the ego-neighbor matching module. In contrast to existing methods that fuse anomalous information from both positive and negative contrastive terms~\cite{liu2021anomaly}, \ourmethod takes a simpler approach. We use the opposite value of the ego-neighbor similarity $c^{(pos)}_i$ as the anomaly score for a node $v_i$, denoted as $s_i = -c^{(pos)}_i$. As discussed in Sec~\ref{subsec:matching}, the similarity $c^{(pos)}_i$ effectively reflects the normality of each node. Hence, by taking the negative value of similarity, we obtain an anomaly score that captures the abnormality of the node. In the inference phase, we can directly discard the negative contrastive terms since the similarity measure already encompasses the information needed to assess the normality of nodes. In contrast to existing contrastive GAD methods that involve negative score calculations and multi-round estimations~\cite{liu2021anomaly, jin2021anemone}, \ourmethod offers improved efficiency and stability during anomaly score estimation.}

\begin{table}[t!]
\centering
\caption{Statistic information of datasets.}
\begin{tabular}{c | c c c c} 
 \toprule
 Dataset & Nodes & Edges & Attributes & Anomalies \\
 \midrule
 Cora & 2,708 & 5,429 & 1,433 & 150 \\
 CiteSeer & 3,327 & 4,732 & 3,703 & 150 \\
 PubMed & 19,717 & 88,648 & 500 & 600 \\
 ACM & 16,484 & 71,980 & 8,337 & 650 \\
 Flickr & 7,575 & 239,738 & 12,407 & 450 \\
 \bottomrule
\end{tabular}
\label{table:datasets}
 \vspace{-3mm}
\end{table}


\subsection{Efficiency Discussion}

\subsubsection{Mini-batch-based Model Training}
\yx{Thanks to the pre-processing of ego and neighbor features, \ourmethod is capable of supporting mini-batch-based optimization, making model training computationally efficient. This allows us to train \ourmethod using batches of ego/neighbor features instead of the entire graph, resulting in reduced memory consumption. Additionally, this feature enables training on edge devices and suggests the potential scalability of \ourmethod.}

\subsubsection{Complexity Analysis}
\yx{We discuss the time complexity of each component in \ourmethod respectively. For the pre-processing module, the complexity of message passing is $\mathcal{O}(kmd)$. Note that this computation can be computed only once in the whole learning procedure. For the ego-neighbor matching module, the complexity of linear mapping is $O(ndd_h)$, where $d_h$ is the hidden dimension of embeddings. The cost of the cosine similarity unit is $O(nd_h)$. To compute the contrastive loss, the time complexity is $O(n)$. The complexity of anomaly scoring is the same as which of the ego-neighbor matching module. To sum up, the pre-processing, training, and testing complexity of \ourmethod are $\mathcal{O}(kmd)$, $O(ndd_hT)$, and $O(ndd_h)$, respectively, where $T$ is the training epoch.}


\section{Experiments}
In this section, we propose our experiment settings and results. We first introduce the datasets, followed by the parameters of training. The experiment results are then presented, including an ablation study and sensitivity analysis. Our method is implemented in PyTorch and is evaluated on a Desktop PC with a Ryzen 5800x CPU, an RTX2070 GPU, and 32GB RAM. The code is available at  \url{https://github.com/CampanulaBells/PREM-GAD}.

\begin{table}[t!]
\centering
\caption{\yx{Hyper-parameters} of our method.}
\resizebox{\columnwidth}{!}{
\begin{tabular}{l | c c c c c } 
 \toprule
 Hyper-parameter & Cora & CiteSeer & PubMed & ACM & Flickr  \\
 \midrule
Embedding dim. $d_h$ & 128 & 128 & 128 & 128 & 128 \\
Prop. step $k$ & 2 & 2 & 2 & 2 & 2 \\
Learning rate &0.0003&0.0003&0.0005&0.0001&0.0005\\
Epoch $T$ &100&100&400&200&1500\\
Trade-off param. $\alpha$ &0.9&0.9&0.6&0.7&0.3\\
Trade-off param. $\gamma$ &0.1&0.1&0.4&0.2&0.4\\
 \bottomrule
\end{tabular}
}
\label{table:param}
 \vspace{-3mm}
\end{table}

\subsection{Experimental Settings}
\subsubsection{Datasets}
The proposed method is evaluated on \bill{five datasets including} four citation network datasets (i.e. Cora, CiteSeer, PubMed, and ACM) and a social network dataset (Flickr). \bill{The specific statistical details are outlined in Table~\ref{table:datasets}. We follow the settings described in prior studies~\cite{ding2019deep, liu2021anomaly} to inject anomalous nodes. Concretely, there are two types of anomaly injection: structural anomalies and attribute anomalies injection. In terms of injecting structural anomalies, we build $\mathit{q}$ cliques of size $\mathit{p}$ by building edges amongst randomly chosen nodes. This results in the formation of $\mathit{pq}$ structural anomalies. For attribute anomalies injection, substitute the feature of the target node with the feature vector of the one exhibiting the highest Euclidean distance in an arbitrarily selected candidate node set. } A total of $\mathit{pq}$ attribute anomalies have been injected for balancing. 

\begin{table}[t!]
\centering
\caption{AUC of \ourmethod and baselines.}
\begin{tabular}{c | c c c c c } 
 \toprule
 Method & Cora & CiteSeer & PubMed & ACM & Flickr \\
 \midrule
DOMINANT & 0.8128 & 0.8197 & 0.7896 & 0.7561 & 0.7436 \\
CoLA & 0.9046 & 0.8894 & 0.9468 & 0.8375 & 0.7616 \\
ANEMONE & \underline{0.9207} & \underline{0.9282} & 0.9534 & 0.8492 & 0.7679 \\
GRADATE & 0.9032 & 0.8904 & 0.9450 &\underline{0.8580} & 0.7522 \\
SL-GAD & 0.9178 & 0.9221 & 0.9627 & 0.8143 & \underline{0.7965} \\
Sub-CR & 0.9023 & 0.9272 & \underline{0.9687} & 0.8060 & 0.7921 \\
 \midrule
\ourmethod & \pmb{0.9510} & \pmb{0.9779} & \pmb{0.9719} & 0.9056 & \pmb{0.8613}\\
\ourmethod-b & 0.9508 & 0.9741 & 0.9714 & \pmb{0.9088} & 0.8561 \\
 \bottomrule
\end{tabular}
\label{table:results}
 \vspace{-3mm}
\end{table}

\begin{figure*}
\centering

\begin{minipage}[t]{0.24\linewidth}
\includegraphics[width=1\linewidth]{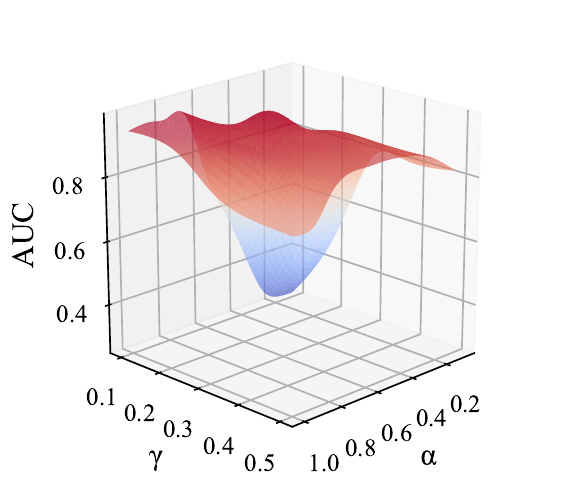}
\subcaption{Cora}
\end{minipage}
\begin{minipage}[t]{0.24\linewidth}
\includegraphics[width=1\linewidth]{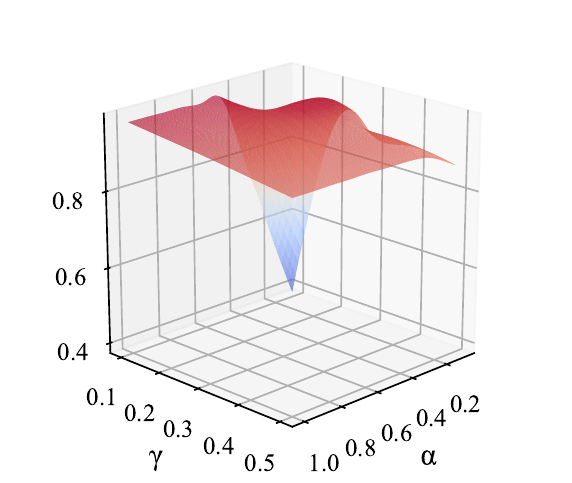}
\subcaption{CiteSeer}
\end{minipage}
\begin{minipage}[t]{0.24\linewidth}
\includegraphics[width=1\linewidth]{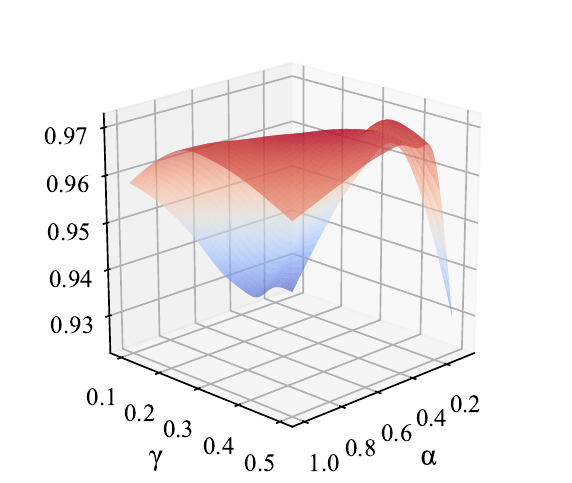}
\subcaption{PubMed}
\end{minipage}
\begin{minipage}[t]{0.24\linewidth}
\includegraphics[width=1\linewidth]{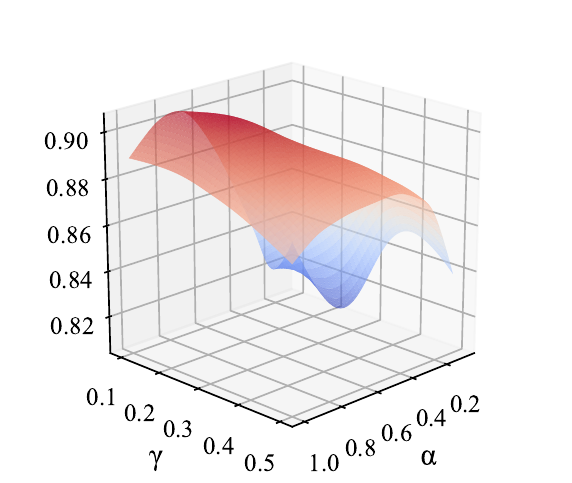}
\subcaption{ACM}
\end{minipage}
\begin{minipage}[t]{0.24\linewidth}
\includegraphics[width=1\linewidth]{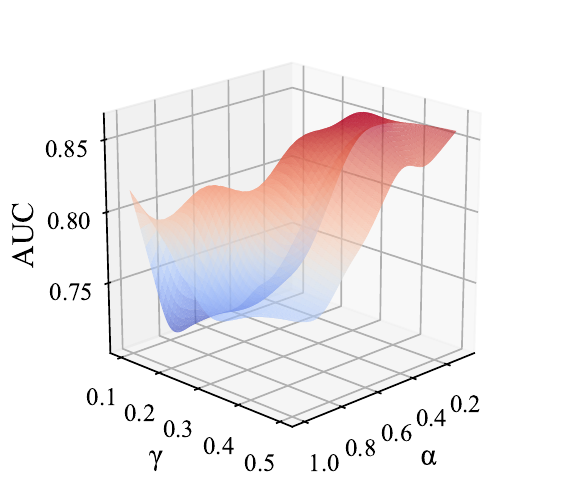}
\subcaption{Flickr}
\end{minipage}
\begin{minipage}[t]{0.24\linewidth}
\centering
\includegraphics[width=0.85\linewidth]{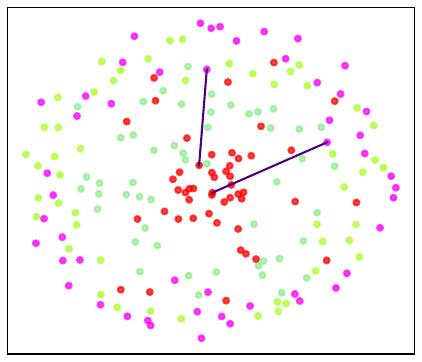}
\subcaption{Features}
\end{minipage}
\begin{minipage}[t]{0.24\linewidth}
\centering
\includegraphics[width=0.85\linewidth]{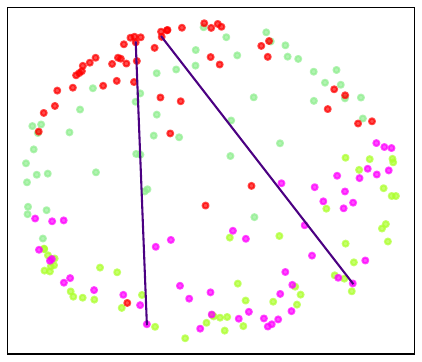}
\subcaption{Embeddings (w Anony.)}
\end{minipage}
\begin{minipage}[t]{0.24\linewidth}
\centering
\includegraphics[width=0.85\linewidth]{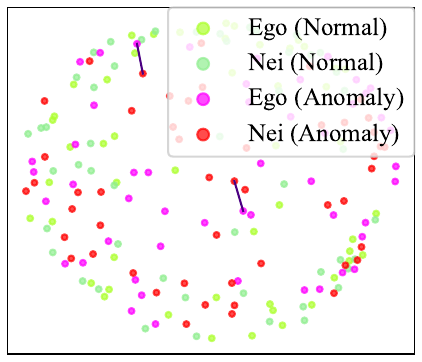}
\subcaption{Embeddings (w/o Anony.)}
\end{minipage}

\centering

\caption{(a)-(e): Sensitivity analysis for balancing factors $\alpha$ and $\gamma$; \yx{(f)-(h) t-SNE visualizations of input features and learned embeddings by \ourmethod (with and without Anonymization) on PubMed dataset.} \bill{In (f)-(h), Ego (Normal/Anomaly) represents ego features for normal and anomaly nodes respectively, while Nei(Nomral/Anomaly) is neighbor feature for normal and anomaly nodes respectively.}}
\label{fig:gridsearch}
\end{figure*}

\begin{table*}[t!]
\centering
\caption{Training and testing time of GAD methods in seconds.}
\begin{tabular}{c | c c | c c | c c | c c | c c} 
 \toprule
 \multirow{2}*{{{Method}}} & \multicolumn{2}{c|}{Cora}& \multicolumn{2}{c|}{CiteSeer}& \multicolumn{2}{c|}{PubMed}& \multicolumn{2}{c|}{ACM}& \multicolumn{2}{c}{Flickr}\\
 \cmidrule(r){2-11}
 & train & test & train & test & train & test & train & test & train & test \\
 \midrule
DOMINANT & \underline{3.9946} & \underline{0.0050} & \underline{8.3463} & \underline{0.0050} & \underline{104} & \underline{0.0100} & \underline{124} & \pmb{0.0090} & \underline{266} & \underline{0.0100} \\
CoLA & 95 & 235 & 180 & 456 & 711 & 1770 & 1639 & 1056 & 712 & 454 \\
ANEMONE & 101 & 247 & 191 & 465 & 732 & 1797 & 1643 & 1047 & 761 & 453 \\
GRADATE & 541 & 319 & 917 & 557 & 3991 & 2392 & 2580 & 1491 & 1234 & 703 \\
SL-GAD & 221 & 546 & 399 & 1007 & 1637 & 4171 & 4585 & 2697 & 2099 & 1262 \\
Sub-CR & 145 & 422 & 243 & 708 & 1104 & 3385 & 720 & 2241 & 1355 & 958 \\
 \midrule
\pmb{\ourmethod} & \pmb{0.5386} & \pmb{0.0007} & \pmb{1.1002} & \pmb{0.0011} & \pmb{4.8098} & \pmb{0.0014} & \pmb{14} & \underline{0.0112} & \pmb{68} & \pmb{0.0071} \\

\ourmethod-b & 5.6882 & 0.0080 & 11 & 0.0190 & 95 & 0.0310 & 214 & 0.1682 & 1028 & 0.1031 \\
 \bottomrule

\end{tabular}
\label{table:runtime}
\end{table*}
\subsubsection{Baselines} 
\yx{In this paper, we compare \ourmethod with six state-of-the-art~(SOTA) baselines, including a generative method (DOMINANT~\cite{ding2019deep}), three contrastive methods (CoLA~\cite{liu2021anomaly}, ANEMONE~\cite{jin2021anemone}, and GADMSL~\cite{duan2022graph}), and two hybrid methods that combines generative and contrastive learning (SL-GAD~\cite{zheng2021generative} and Sub-CR~\cite{zhang2022reconstruction}).}

\subsubsection{Evaluation metric}

We utilize ROC-AUC as our evaluation metric to compare the results. The ROC curve is the plot of the true positive rate and false positive rate under different thresholds. AUC is the area under the ROC curve which is capped by 1. The larger the AUC, the better the classifier. We also record the time taken to train and evaluate methods in seconds to compare the efficiency of methods.

\subsubsection{Implementation Details} 
\bill{We present the parameter setting of \ourmethod for the five datasets on Table~\ref{table:param}.}
\bill{We \yx{set the number of propagation steps $k$ to 2} and set the embedding dimension $d_h$ to 128 for all datasets}. The learning rate of \ourmethod for Cora, CiteSeer, PubMed, ACM, and Flickr are 3e-4, 3e-4, 5e-4, 1e-4, 5e-4, and the epochs are 100, 100, 400, 200, 1500, respectively, to get better performance on datasets with various size. For balance factors $\alpha$ and $\gamma$, we \bill{use} grid search with search space $[0.1, 0.2, \dots, 1.0]$ for $\alpha$ and $[0.1, 0.2, \dots, 0.5]$ for $\gamma$. \yx{We also consider \ourmethod with a mini-batch size of 300 (denoted as ``\ourmethod-b'') for comparison.} Our method has been evaluated for 10 times with different random seeds to report the average results. For baselines, we follow the settings in their paper \bill{to obtain the reported results. } 

\begin{table*}[t!]
\centering
\caption{Maximum GPU memory usage of GAD methods in MB.}
\begin{tabular}{c | c c | c c | c c | c c | c c} 
 \toprule
 \multirow{2}*{{{Method}}} & \multicolumn{2}{c|}{Cora}& \multicolumn{2}{c|}{CiteSeer}& \multicolumn{2}{c|}{PubMed}& \multicolumn{2}{c|}{ACM}& \multicolumn{2}{c}{Flickr}\\
 \cmidrule(r){2-11}
 & train & test & train & test & train & test & train & test & train & test \\
 \midrule
DOMINANT & 282 & 282 & 572 & 572 & 5943 & 5943 & 6051 & 6051 & 4399 & 4401 \\
CoLA & \underline{60} & \underline{60} & \underline{137} & \underline{135} & 1561 & \underline{1528} & 2088 & \underline{1665} & 918 & \underline{717} \\
ANEMONE & 62 & 62 & 139 & 139 & \underline{1533} & 1529 & \underline{1673} & 1673 & \underline{729} & 729 \\
GRADATE & 95 & 90 & 185 & 182 & 3022 & 3015 & 2713 & 2713 & 951 & 949 \\
SL-GAD & 164 & 129 & 415 & 330 & 1600 & 1591 & 2691 & 2588 & 1784 & 1541 \\
Sub-CR & 128 & 118 & 276 & 263 & 3083 & 3056 & 3652 & 3309 & 1491 & 1405 \\
 \midrule
\ourmethod & 110 & 54 & 347 & 161 & 294 & 146 & 3711 & 1629 & 2495 & 1103 \\
\ourmethod-b & \pmb{16} & \pmb{9} & \pmb{44} & \pmb{24} & \pmb{6} & \pmb{3} & \pmb{97} & \pmb{52} & \pmb{138} & \pmb{75} \\

\bottomrule
\end{tabular}
\label{table:memory}
\end{table*}

\begin{table}[t!]
\centering
\caption{Results of ablation analysis.}
\resizebox{\columnwidth}{!}{
\begin{tabular}{l | c c c c c | c } 
 \toprule
 Variant & Cora & CiteSeer & PubMed & ACM & Flickr & Average\\
 \midrule
w/o nei. neg. & 0.7369  & 0.7304  & 0.8848  & 0.6131  & 0.7565 & 0.7393\\
w/o ego neg. & 0.9310  & \pmb{0.9782}  & 0.9210  & 0.8730  & 0.7048 & 0.8761\\
w/o Anony. & \pmb{0.9541}  & 0.9592  & 0.8811  &  \pmb{0.9109}  & 0.8536 & 0.9109\\
 \midrule
\ourmethod & 0.9510 & 0.9779 & \pmb{0.9719} &0.9056 & \pmb{0.8613} & \pmb{0.9328}\\
 \bottomrule
\end{tabular}
}
\label{table:ablation}
\end{table}

\subsection{Comparative Results}
In this subsection, we evaluate our proposed \ourmethod by comparing its performances and training and evaluation time to other baseline methods. The AUC of \ourmethod and baselines are summarized in Table~\ref{table:results}, and the training and testing time are listed in Table~\ref{table:runtime}. Fig.~\ref{fig:ROC} shows ROC curves. 

\subsubsection{Anomaly Detection Performance}
\bill{From the data presented in Table~\ref{table:results}, it is evident that \ourmethod achieves the best performance across all datasets, markedly outperforming baselines. This accomplishment is a testament to its robustness and the efficacy of its design. For example, \ourmethod achieves the highest performance in the CiteSeer dataset, where it achieves an AUC of 0.9779. This result is not only the best in the group but it's also noticeably higher than the other methods, i.e., leading by around 5\%, illustrating the effectiveness of \ourmethod. The PubMed and ACM datasets further underscore \ourmethod's dominance. It not only processes these large datasets efficiently but also manages to deliver superior results with an AUC of 0.9719 and 0.9056 respectively.} 
\yx{We can also witness that the mini-batch-based training would not lead to performance degradation for \ourmethod. }

\subsubsection{Anomaly Detection Efficiency}
\bill{Table~\ref{table:runtime} provides a detailed comparison of the training and testing times of \ourmethod and various baseline methods across five datasets. Impressively, \ourmethod far surpasses all other methods in terms of efficiency. On average, it often operates at a pace 10 times swifter than its closest competitor , DOMINANT in training, and 100 times faster than other contrastive learning based methods. Significantly, in testing, this efficiency gap between our method and other contrastive learning-based methods widens, e.g., \ourmethod is 100,000 times faster than ANEMONE, as computing negative ego-neighbor similarity is much more efficient than subgraph sampling and multiple round evaluation, which are widely used in other contrastive learning based methods~\cite{liu2021anomaly, jin2021anemone, duan2022graph}. This speed advantage is attained without compromising memory efficiency or performance. While the fastest baseline DOMINANT demonstrates the ability to manage larger datasets, it does so at the expense of substantial memory consumption. For instance, with the PubMed dataset, DOMINANT consumes roughly 5943 MB of memory in training, which is about 20 times more than \ourmethod's mere 294. \yx{Moreover, the mini-batch version \ourmethod-b has extremely small memory requirements across all datasets, enlightening the potential of running \ourmethod on edge devices with limited memory.} The efficiency and effectiveness of \ourmethod stem from its simplicity as we only use a discriminator for anomaly score calculation and most of the heavy computation can be done and reused in the pre-processing module. It avoids complex objectives and modules that other baselines do, focusing instead on efficiency and precision. This strategy minimizes computational costs and reduces processing time while achieving SOTA performance.}



\subsection{Ablation Study}
In this subsection, we study the effect of negative contrast pairs and anonymization. The results are listed in Table~\ref{table:ablation}. Overall, we can observe that \ourmethod achieves the best results when all components are included in terms of the average result. Disabling the neighbor negative term results in the AUC drop to the bottom, which illustrates that this term plays a primary role in the model training. While removing the ego negative term slightly increases the AUC on CiteSeer, it has a significant impact on large datasets such as PubMed and Flickr. We can also find that using the two modules can improve anomaly detection performance. \bill{When examining the impact of anonymization, a comparison between Fig.~\ref{fig:gridsearch}~(g) and (h) reveals a significant reduction in the distance between ego features and node features of anomalous nodes when anonymization is not applied. As a result, anomaly nodes become more difficult to be distinguished from their neighborhood and, thus lead to the degradation of model performance.}

\subsection{Sensitivity Analysis}
To investigate the effect of the trade-off parameters $\alpha$ and $\gamma$, we perform a grid search with stride=0.1 for all parameters. As shown in Fig.~\ref{fig:gridsearch}~(a)-(e), a similar trend is observed among citation networks. \ourmethod achieves the highest AUC scores when $\alpha$ and $\gamma$ are in equilibrium, while its performance decreases when either $\alpha$ or $\gamma$ is approaches extreme values. When both $\alpha$ and $\gamma$ are 0.1, the AUC drops significantly, indicating the collapse problem. The AUC on the social network Flickr is more sensitive to changes in hyper-parameters, peaking at $\alpha=0.3$ and $\gamma=0.4$. To conclude, the trade-off between neighbor negative and ego negative depends on the properties of datasets, which is consistent with previous findings~\cite{duan2022graph, zheng2021generative, liu2022dagad}. While hyper-parameter turning becomes expensive for large datasets, it is possible to quickly find the most appropriate trade-off parameters of \ourmethod for large datasets with limited computational resources due to the highly efficient and scalable framework and training procedure. \looseness-1

\section{Conclusion}

This paper introduces \ourmethod, a simple yet powerful graph anomaly detection method. By leveraging a pre-processing module to pre-compute neighbor features, \ourmethod eliminates the need for train-time message passing, resulting in high training efficiency. The prediction of anomaly scores is achieved through an ego-neighbor matching module that utilizes cosine similarity between ego and neighbor embeddings, leading to fast and accurate abnormality estimation. Extensive experiments demonstrate that \ourmethod significantly outperforms state-of-the-art methods by a large margin, while also exhibiting high running efficiency and low memory consumption. In future work, we plan to explore the extension of \ourmethod to address the anomaly detection problem in complex graph types, such as dynamic graphs, heterogeneous graphs, and heterophilic graphs. \looseness-2

\bibliographystyle{plain}
\bibliography{refs}

\end{document}